\documentclass[lettersize,journal,comsoc]{IEEEtran}
\usepackage{amsmath,amsfonts}
\usepackage{algorithmic}
\usepackage{algorithm}
\usepackage{array}
\usepackage[caption=false,font=normalsize,labelfont=sf,textfont=sf]{subfig}
\usepackage{textcomp}
\usepackage{stfloats}
\usepackage{url}
\usepackage{verbatim}
\usepackage{graphicx}
\usepackage{cite}
\usepackage[hidelinks]{hyperref}
\usepackage{subfig}

\begin{document}

\title{Autoencoder Models for Point Cloud Environmental Synthesis from WiFi Channel State Information: A Preliminary Study}

\author{Daniele Pannone,~\IEEEmembership{Member,~IEEE,} Danilo Avola,~\IEEEmembership{Member,~IEEE}
\thanks{Daniele Pannone and Danilo Avola are with the Department of Computer Science, Sapienza University of Rome, Via Salaria 113, 00198, Rome, Italy. Emails: \{pannone,avola\}@di.uniroma1.it}
\thanks{Manuscript received April 19, 2021; revised August 16, 2021.}}

\markboth{Journal of XXX}%
{Pannone et al.}


\maketitle

\begin{abstract}
This paper introduces a deep learning framework for generating point clouds from WiFi Channel State Information data. We employ a two-stage autoencoder approach: a PointNet autoencoder with convolutional layers for point cloud generation, and a Convolutional Neural Network autoencoder to map CSI data to a matching latent space. By aligning these latent spaces, our method enables accurate environmental point cloud reconstruction from WiFi data. Experimental results validate the effectiveness of our approach, highlighting its potential for wireless sensing and environmental mapping applications.
\end{abstract}


\section{Introduction}
\IEEEPARstart{T}{he} proliferation of wireless communication technologies has led to an increased interest in using WiFi signals for various sensing applications. Among these, Channel State Information (CSI) data from WiFi signals provides rich information about the environment, making it a valuable resource for tasks such as indoor localization\cite{zhao2018through,adib2015multi}, activity recognition\cite{hasmath2020device,wang2014eyes}, and environmental mapping\cite{adib2013see,donarski2020environment}. However, translating CSI data into meaningful spatial representations, such as point clouds, remains a challenging problem due to the complex and high-dimensional nature of the data.

Point clouds are a fundamental data structure in 3D computer vision, offering detailed spatial information that can be used for a wide range of applications, including object recognition\cite{charles2017pointnet}, scene understanding\cite{zhou2018voxelnet}, and robotics\cite{gupta2015aligning}. Traditional methods for generating point clouds often rely on specialized hardware like LiDAR or depth cameras, which can be expensive and limited in their deployment. In contrast, WiFi signals are ubiquitous and can be captured using commodity hardware, making them an attractive alternative for point cloud generation.

In this paper, we propose a novel deep learning framework for generating point clouds from WiFi CSI data. Our approach leverages a two-stage autoencoder architecture. The first stage employs a PointNet\cite{charles2017pointnet} autoencoder with convolutional layers to learn a latent space representation of point clouds. The second stage uses a Convolutional Neural Network (CNN)\cite{lecun2015deep} autoencoder to map CSI data to a latent space that matches the PointNet autoencoder latent space. By aligning these latent spaces, our method allows the generation of accurate point clouds of environment directly from WiFi CSI data.

The contributions of this work are threefold:
\begin{enumerate}
    \item We introduce a deep learning framework that combines a convolutional PointNet autoencoder with a CNN autoencoder for point cloud generation from WiFi CSI data;
    \item We demonstrate the effectiveness of our approach through experimental results, showing that our method can accurately reconstruct point clouds from CSI data;
    \item We discuss the potential applications of our framework in wireless sensing and environmental mapping, highlighting its advantages over traditional point cloud generation methods.
\end{enumerate}

The remainder of this paper is organized as follows: Section \ref{sec:background_pc} provides a background on point cloud generation, while Section \ref{sec:background_csi} provides a background on CSI. Section \ref{sec:method} describes our proposed deep learning framework in detail. Section \ref{sec:experiments} presents our experimental setup and results. Finally, Section \ref{sec:conclusion} concludes the paper.

\section{Background on Point Clouds}
\label{sec:background_pc}
Point clouds are a fundamental data structure in 3D computer vision and geometry processing, representing a set of data points in a three-dimensional space. Formally, a point cloud $\mathcal{P}$ can be defined as a set of $N$ points, where each point $\mathbf{p}_i$ is a vector in $\mathbb{R}^3$ representing its spatial coordinates:

\begin{equation}
\mathcal{P} = \{ \mathbf{p}_i \mid \mathbf{p}_i \in \mathbb{R}^3, i = 1, 2, \ldots, N \}.
\end{equation}

Each point $ \mathbf{p}_i $ typically includes spatial coordinates $ (x_i, y_i, z_i) $ and may also include additional attributes such as color, intensity, or normal vectors. These attributes can be represented as an extended vector $ \mathbf{a}_i \in \mathbb{R}^m $, where $ m $ is the dimension of the attribute space. Thus, an augmented point cloud can be defined as:
\begin{equation}
\mathcal{P}' = \{ (\mathbf{p}_i, \mathbf{a}_i) \mid \mathbf{p}_i \in \mathbb{R}^3, \mathbf{a}_i \in \mathbb{R}^m, i = 1, 2, \ldots, N \}  .  
\end{equation}

A point cloud can be generated by exploiting different approaches and sensors. LiDAR (Light Detection and Ranging) sensors emit laser pulses and measure the time it takes for the pulses to reflect back, generating highly accurate and dense point clouds. A similar approach is used by RGB-D cameras, which capture both color (RGB) and depth (D) information, allowing for the generation of point clouds by projecting the depth information into 3D space. Finally, stereo vision and photogrammetry use standard RGB images for building the point cloud. The former uses two or more cameras to capture images from different viewpoints. By finding corresponding points in the images and triangulating their positions, stereo vision systems can generate point clouds, while the latter uses multiple photographs of an object or scene from different angles and using computer vision algorithms to reconstruct the 3D geometry.
\begin{figure*}[t]
    \centering
    \includegraphics[width=0.7\linewidth]{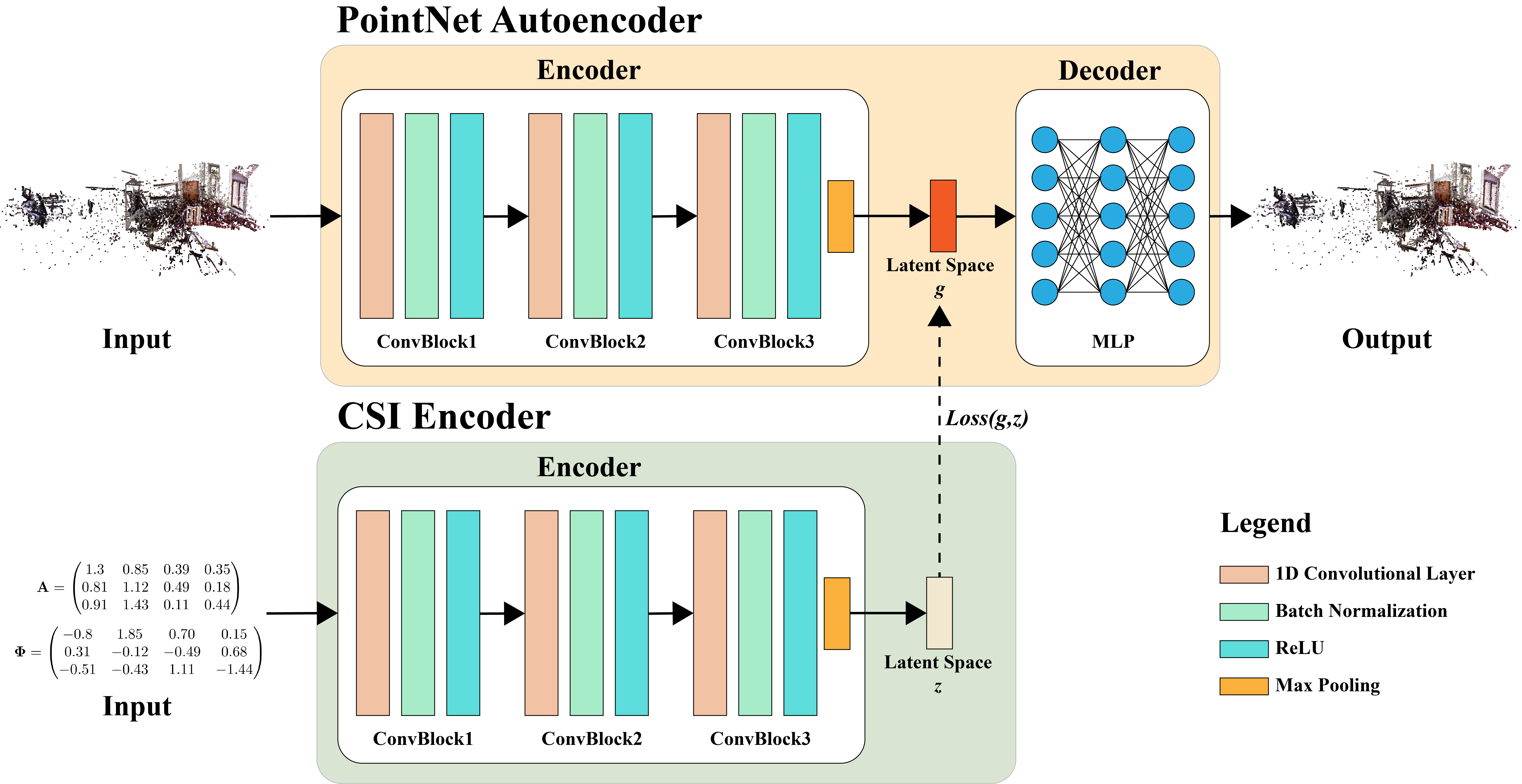}
    \caption{Architecture of the Proposed Framework for Generating Point Clouds from WiFi CSI Data. The proposed framework consists of two main components: a PointNet autoencoder with convolutional layers and a CSI encoder. The PointNet autoencoder is trained to generate point clouds from a latent space representation, while the CSI encoder is trained to map CSI data to a latent space that matches the PointNet autoencoder latent space. The PointNet autoencoder processes the input point cloud through multiple 1D convolutional layers and max pooling to extract hierarchical features, which are then aggregated into a global feature vector serving as the latent space representation. The decoder uses fully connected layers to reconstruct the point cloud from this latent space. The CSI encoder processes the amplitude and phase components extracted from CSI, through 1D convolutional layers to extract features and map them to a matching latent space. The alignment of these latent spaces enables the generation of accurate point clouds directly from WiFi CSI data.}
    \label{fig:model}
\end{figure*}
\section{Background on CSI}
\label{sec:background_csi}
CSI is a critical parameter in wireless communication systems, providing detailed information about the communication channel between the transmitter and the receiver. CSI captures the effects of scattering, fading, and power decay with distance, offering insights into the propagation environment. Mathematically, CSI can be represented as a complex-valued matrix that describes the channel characteristics across multiple subcarriers and antennas. Consider a Multiple-Input Multiple-Output (MIMO) Orthogonal Frequency-Division Multiplexing (OFDM) system with $ N_t $ transmit antennas and $ N_r $ receive antennas. The channel response for the $ k $-th subcarrier can be represented as a complex matrix $ \mathbf{H}_k \in \mathbb{C}^{N_r \times N_t} $, where each element $ h_{ij}^k $ denotes the channel gain from the $ j $-th transmit antenna to the $ i $-th receive antenna for the $ k $-th subcarrier. The CSI matrix $ \mathbf{H}_k $ can be expressed as:

\begin{equation}
\mathbf{H}_k = \begin{pmatrix}
h_{11}^k & h_{12}^k & \cdots & h_{1N_t}^k \\
h_{21}^k & h_{22}^k & \cdots & h_{2N_t}^k \\
\vdots & \vdots & \ddots & \vdots \\
h_{N_r1}^k & h_{N_r2}^k & \cdots & h_{N_rN_t}^k \\
\end{pmatrix},
\end{equation}

\noindent while the received signal $ \mathbf{y}_k \in \mathbb{C}^{N_r \times 1} $ for the $ k $-th subcarrier can be modeled as:

\begin{equation}
\mathbf{y}_k = \mathbf{H}_k \mathbf{x}_k + \mathbf{n}_k,
\end{equation}

\noindent where $ \mathbf{x}_k \in \mathbb{C}^{N_t \times 1} $ is the transmitted signal vector, and $ \mathbf{n}_k \in \mathbb{C}^{N_r \times 1} $ is the additive white Gaussian noise (AWGN) vector with zero mean and variance $ \sigma^2 $.

Regarding the extraction of CSI, CSI is typically estimated using pilot signals known to both the transmitter and the receiver. The estimation process involves transmitting known pilot symbols $ \mathbf{p}_k $ and measuring the received signals $ \mathbf{r}_k $. The estimated CSI matrix $ \hat{\mathbf{H}}_k $ for the $ k $-th subcarrier can be obtained using techniques such as Least Squares (LS) or Minimum Mean Square Error (MMSE) estimation. For LS estimation, the CSI matrix is given by:

\begin{equation}
\hat{\mathbf{H}}_k = \mathbf{r}_k \mathbf{p}_k^\dagger (\mathbf{p}_k \mathbf{p}_k^\dagger)^{-1},
\end{equation}

\noindent where $ \mathbf{p}_k^\dagger $ denotes the Hermitian transpose of the pilot vector $ \mathbf{p}_k $.

\section{Proposed Method}
\label{sec:method}

In this section, we present our proposed method for generating point clouds from CSI data. The method consists of two main components: a PointNet autoencoder with convolutional layers and a CSI encoder using 1D convolutions. The PointNet autoencoder is trained to generate point clouds from a latent space representation, while the CSI encoder maps CSI data to a latent space that matches the PointNet autoencoder latent space. The graphical representation of our method is shown in Fig. \ref{fig:model}, and in the following sections, the several components of our method are explained in detail.

\subsection{PointNet Autoencoder}
The PointNet autoencoder is designed to process point cloud data and generate a latent space representation. The architecture includes an encoder that extracts features from the input point cloud and a decoder that reconstructs the point cloud from the latent space representation.

\subsubsection{Encoder}
The encoder consists of several layers that process the input point cloud to extract hierarchical features. The input layer gets the input point cloud $ \mathbf{X} \in \mathbb{R}^{N \times 3} $, which consists of $ N $ points, each with 3 coordinates (x, y, z). Subsequently, several 1D convolutional layers are applied to extract local features from the input point cloud. Let $ \mathbf{F}^{(l)} $ be the feature matrix after the $ l $-th convolutional layer. For each layer $ l $, we have that $ \mathbf{F}^{(l)} = \text{Conv1D}(\mathbf{F}^{(l-1)}) $. Each convolutional layer is then followed by a batch normalization layer and an activation function, which in our case is the ReLU\cite{agarap2019deeplearningusingrectified}. These 3 layers together create a convolutional block, and in the proposed encoder, 3 convolutional blocks have been used. After the last convolutional block, a max-pooling layer is used to aggregate local features into a global feature vector. Let $ L $ be the final layer of the encoder. The global feature vector can be defined as $ \mathbf{g} = \text{MaxPool}(\mathbf{F}^{(L)}) $. To resume, the encoder can be formally represented as:

\begin{align}
\mathbf{F}^{(0)} = \mathbf{X} \\
\mathbf{F}^{(l)} = \text{Conv1D}(\mathbf{F}^{(l-1)}) \quad \text{for} \quad l \in [1,3) \\
\mathbf{g} = \text{MaxPool}(\mathbf{F}^{(L)}),
\end{align}

\noindent where $ g $ is the generated latent space.

\subsubsection{Decoder}
The decoder reconstructs the point cloud from the latent space representation generated with the encoder, and in our PointNet autoencoder, the decoder consists of fully connected layers. As for the encoder, also the decoder consists of 3 layers. Let $ \mathbf{h}^{(m)} $ be the hidden representation after the $ m $-th fully connected layer. We have that, for each layer $ m $, $ \mathbf{h}^{(m)} = \text{FC}(\mathbf{h}^{(m-1)}) $, where $ \text{FC} $ denotes a fully connected layer. More formally, the decoder can be resumed as:

\begin{align}
\mathbf{h}^{(0)} = \mathbf{g} \\
\mathbf{h}^{(m)} = \text{FC}(\mathbf{h}^{(m-1)}) \quad \text{for} \quad m \in [1,3) \\
\mathbf{\hat{X}} = \mathbf{h}^{(M)} ,
\end{align}

\noindent where $ \mathbf{\hat{X}} $ is the reconstructed point cloud.

\subsection{CSI Encoder}
The CSI encoder is designed to map the input CSI data to a latent space that matches the PointNet autoencoder latent space. The architecture includes 1D convolutional layers that process the input CSI data to extract features. Let us define the raw CSI data as $ \mathbf{H} \in \mathbb{C}^{M \times K} $, where $ M $ is the number of subcarriers and $ K $ is the number of antennas. The amplitude $ \mathbf{A} $ and phase $ \mathbf{\Phi} $ components can be extracted as follows:

\begin{align}
\mathbf{A} = |\mathbf{H}| \\
\mathbf{\Phi} = \angle \mathbf{H}.
\end{align}

Hence, the input for the first layer consists in a matrix  $ \mathbf{C} \in \mathbb{R}^{M \times 2K} $, where $ M $ is the number of subcarriers, and $ 2K $ are the features extracted for $ K $ antennas, namely,  $ \mathbf{A} $ and $ \mathbf{\Phi} $.

As for the PointNet encoder, also for the CSI encoder we use 3-1D convolutional blocks. Let $ \mathbf{H}^{(l)} $ be the feature matrix after the $ l $-th convolutional layer. For each layer $ l $, we have that $ \mathbf{H}^{(l)} = \text{Conv1D}(\mathbf{H}^{(l-1)}) $. The final feature matrix $ \mathbf{H}^{(L)} $ serves as the latent space representation for the CSI data. The CSI encoder can be formally resumed as follows:

\begin{align}
\mathbf{H}^{(0)} = \mathbf{C} \\
\mathbf{H}^{(l)} = \text{Conv1D}(\mathbf{H}^{(l-1)}) \quad \text{for} \quad l \in [1,3)] \\
\mathbf{z} = \mathbf{H}^{(L)} ,
\end{align}

\noindent where $ z $ is the generated latent space.

\subsection{Loss Functions}
For the PointNet autoencoder, the Chamfer Distance is used as loss function. Such distance is a commonly used metric for comparing two point clouds, as it measures the average distance between each point in the starting point cloud and its nearest neighbor in the destination point cloud. Given two point clouds $ \mathcal{P} = \{ \mathbf{p}_i \mid \mathbf{p}_i \in \mathbb{R}^3, i = 1, 2, \ldots, N \} $ and $ \mathcal{Q} = \{ \mathbf{q}_j \mid \mathbf{q}_j \in \mathbb{R}^3, j = 1, 2, \ldots, M \} $, the Chamfer Distance $ \mathcal{D}_{\text{CD}} $ is defined as:
\begin{equation}
\resizebox{0.43\textwidth}{!}{$\mathcal{D}_{\text{CD}}(\mathcal{P}, \mathcal{Q}) = \frac{1}{N} \sum_{i=1}^{N} \min_{j} \| \mathbf{p}_i - \mathbf{q}_j \|^2 + \frac{1}{M} \sum_{j=1}^{M} \min_{i} \| \mathbf{q}_j - \mathbf{p}_i \|^2,$}
\end{equation}

\noindent where $ \mathbf{p}_i $ and $ \mathbf{q}_j $ are points in the point clouds $ \mathcal{P} $ and $ \mathcal{Q} $, respectively, $ \| \mathbf{p}_i - \mathbf{q}_j \| $ denotes the Euclidean distance between points $ \mathbf{p}_i $ and $ \mathbf{q}_j $, $ \min_{j} \| \mathbf{p}_i - \mathbf{q}_j \|^2 $ finds the nearest neighbor of $ \mathbf{p}_i $ in $ \mathcal{Q} $, and $ \min_{i} \| \mathbf{q}_j - \mathbf{p}_i \|^2 $ finds the nearest neighbor of $ \mathbf{q}_j $ in $ \mathcal{P} $

Instead, for the CSI encoder, the Mean Squared Error (MSE) loss function has been used. In detail, the loss function is computed between the latent space representation of the PointNet autoencoder $ \mathbf{g} $ and the latent space generated from the CSI encoder $ \mathbf{z} $:

\begin{equation}
\mathcal{L}_{\text{CSI}} = \| \mathbf{g} - \mathbf{z} \|^2.
\end{equation}

The second loss function allows mapping the latent space generated from CSI data to the latent space generated from point cloud data.

\section{Experiments}
\label{sec:experiments}
\begin{figure*}[t]
       \subfloat[]{\includegraphics[width=0.3\textwidth]{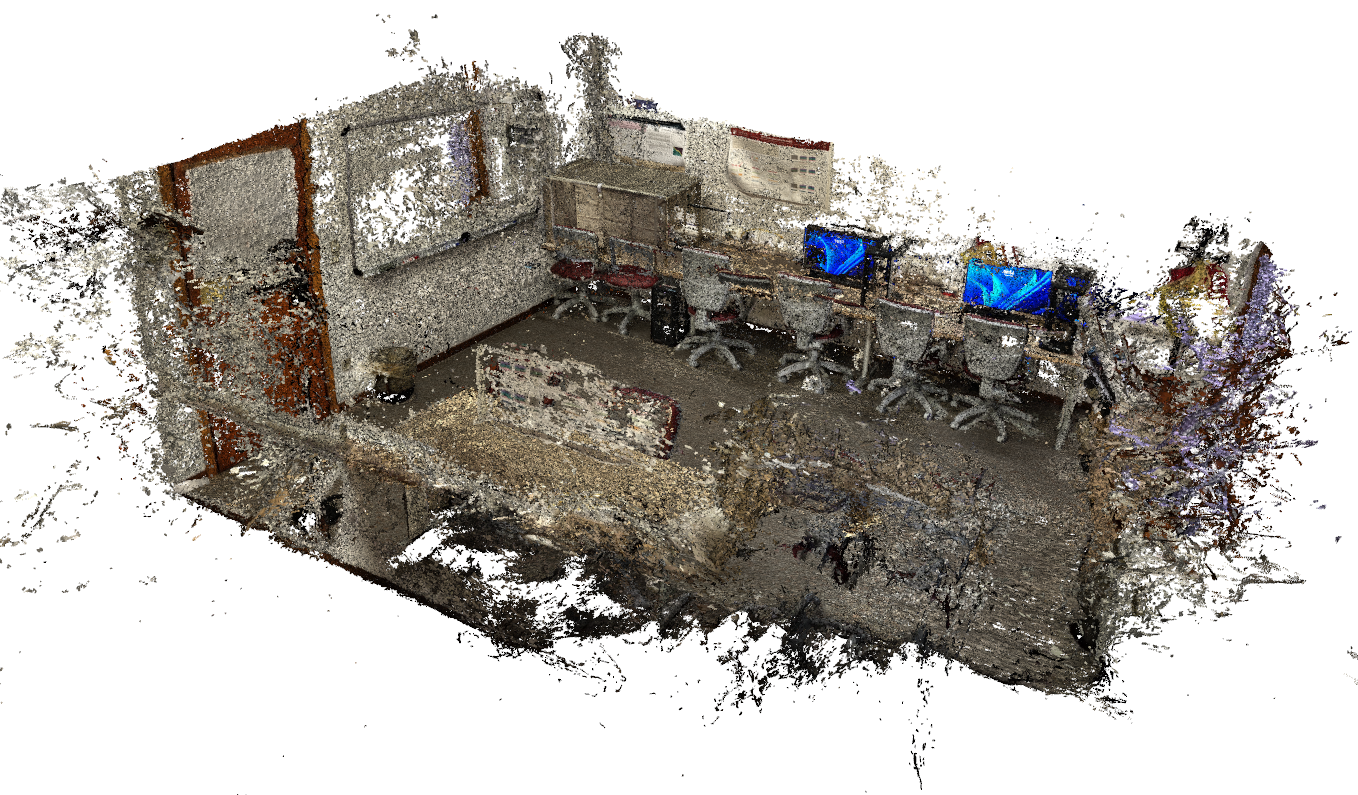}}
       \hfill
       \subfloat[]{\includegraphics[width=0.3\textwidth]{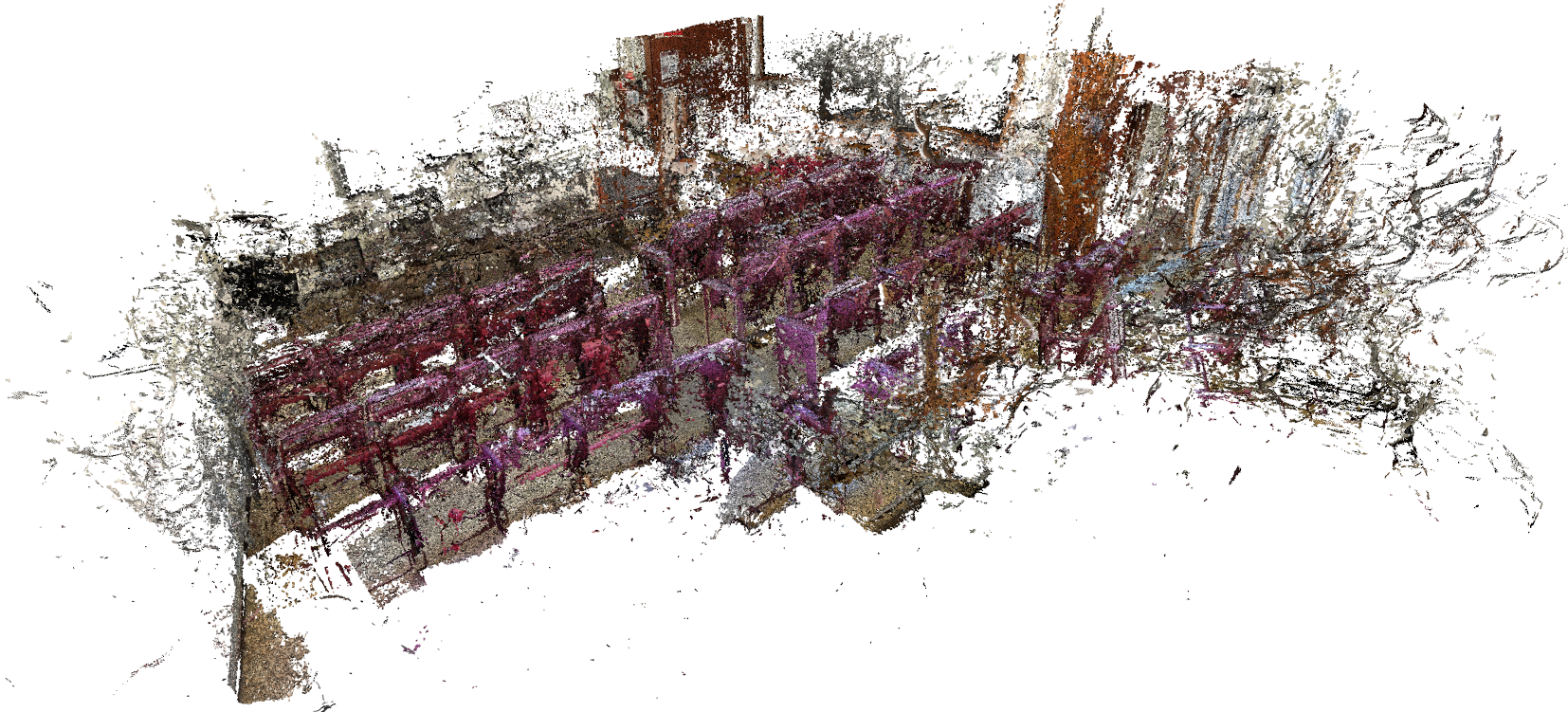}}
       \hfill
       \subfloat[]{\includegraphics[width=0.3\textwidth]{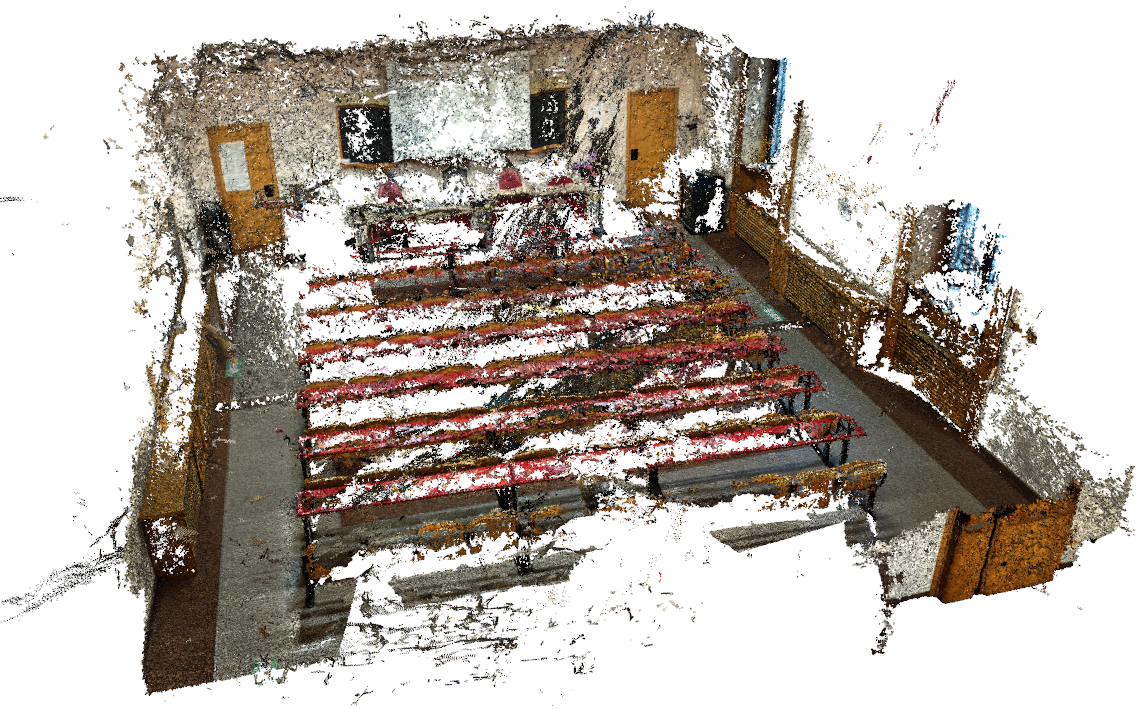}}
        \caption{The point clouds generated from the images of acquired room. These point clouds have been used for training the CSI Encoder and to subsequently test it.}
        \label{fig:original_pointclouds}
\end{figure*}

This section presents the experiments conducted to evaluate the effectiveness of our proposed method for generating point clouds from WiFi CSI data.

\subsection{Dataset}
Due to the duality of our data, we used two different datasets for training the different parts of the proposed model. The dataset used for training the PointNet autoencoder is the LiDAR-Net\cite{guo2024lidarnet}. LiDAR-Net is a new dataset of real-scanned indoor point clouds, featuring nearly 3.6 billion precisely annotated points covering 30,000 $m^2$. It includes three common daily environments: learning, working, and living scenes. The dataset is notable for its non-uniform point distribution, which enhance the training of deep learning models. Regarding the CSI dataset, the latter contains both point cloud and CSI data of 3 rooms acquired by us. The acquisitions are made with 2 Espressif ESP32\footnote{\url{https://www.espressif.com/en/products/socs/esp32}} devices, placed in opposite corners of the acquired rooms. Each room has been acquired 10 times, with a frequency of acquisition of 100 packets per seconds. The point cloud associated with the CSI data was generated from RGB images using COLMAP\cite{schoenberger2016sfm}, and the resulting point clouds are illustrated in Fig. \ref{fig:original_pointclouds}.

\subsection{Model Training and Implementation}
The model training was conducted in two stages. In the first stage, the PointNet model was trained on the LiDAR-Net dataset to establish a robust latent space for point cloud generation. The second stage involved training the CSI Encoder to align its latent space with the one learned by the PointNet model. For both the training stages, the AdamW optimizer was used with a learning rate of $0.001$. The used deep learning framework was Pytorch, and the models were trained on an NVidia RTX 4090 GPU with 24 GB of RAM.

\subsection{Results}
\begin{table*}[t]
    \centering
    \caption{Baseline results obtained in our experiments.}
    \begin{tabular}{lcccc}
        \hline
        \hline
        \textbf{Method} & \textbf{Mean CD $\downarrow$} & \textbf{Std Dev} & \textbf{Mean EMD $\downarrow$} & \textbf{Std Dev} \\
        Direct CSI-to-PointCloud Regression  & 0.325  & 0.420 & 0.412  & 0.510  \\
        Proposed method & \textbf{0.083}  & 0.029 & \textbf{0.175}  & 0.033  \\
        \hline
        \hline
    \end{tabular}
    \label{tab:results}
\end{table*}

In our experiments, we compared the proposed method with a baseline approach, i.e., the Direct CSI-to-PointCloud Regression. This approach consists in a model trained to map CSI directly to point clouds without latent space alignment.

\begin{figure*}[t]
       \subfloat[]{\includegraphics[width=0.3\textwidth]{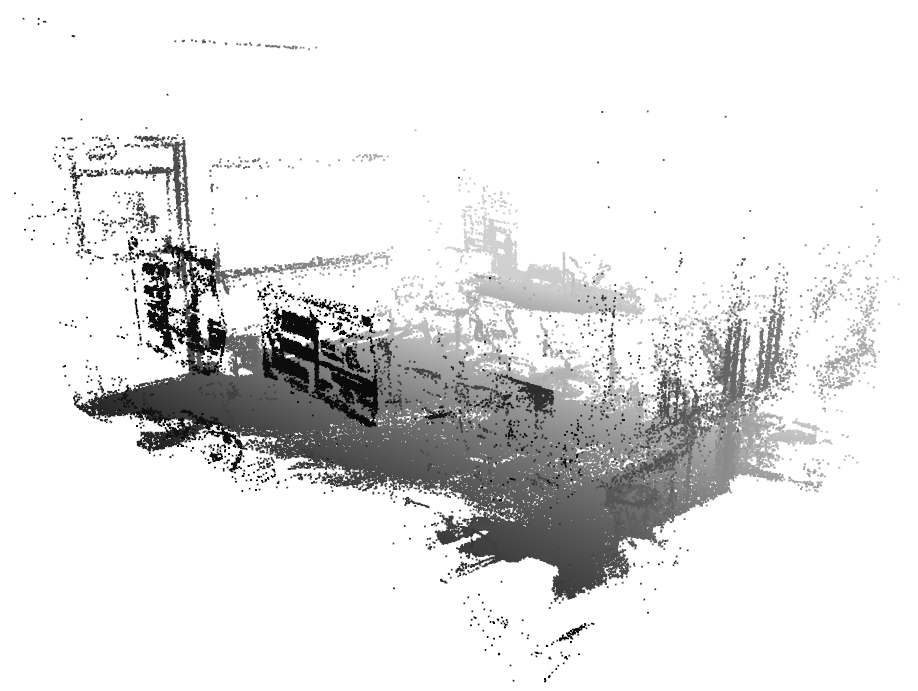}}
       \hfill
       \subfloat[]{\includegraphics[width=0.3\textwidth]{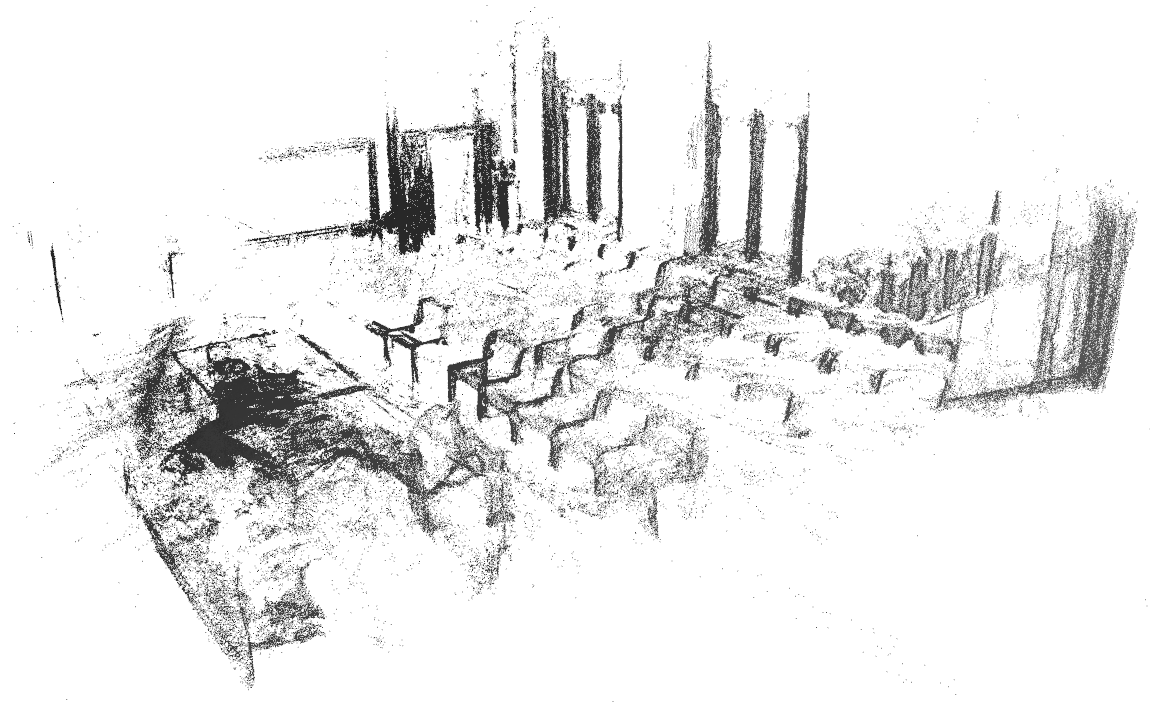}}
       \hfill
       \subfloat[]{\includegraphics[width=0.3\textwidth]{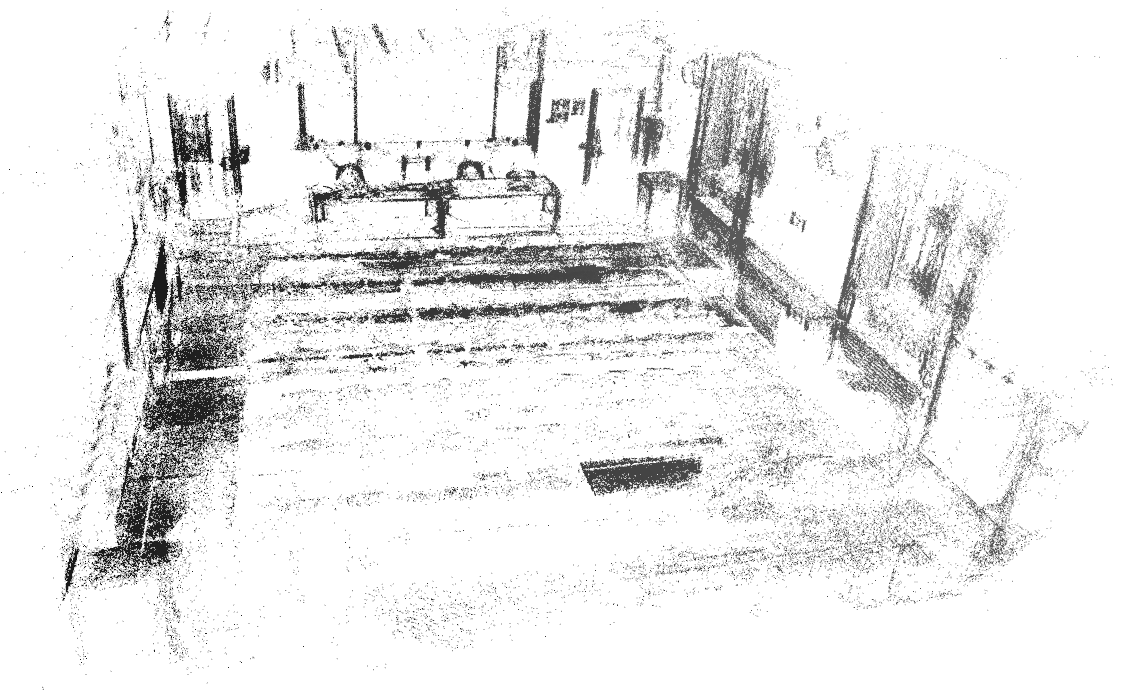}}
        \caption{The point clouds generated from the CSI data of a) the first, b) the second, and c) the third room, respectively.}
        \label{fig:rec_pointclouds}
\end{figure*}

For comparing the reconstructed point clouds with the ground truth, two well-established metrics have been used: the \textbf{Chamfer Distance (CD)}, which measures average closest-point distances, and the \textbf{Earth Mover's Distance (EMD)}, which quantifies the minimal cost to transform one point set into another.
Table \ref{tab:results} presents the result obtained with the proposed method, and compares them with the baseline approach. This comparison reveals significant differences in performance, generalization capabilities, and learning stability. The direct regression model attempts to infer geometric layouts from CSI data in an end-to-end approach. However, CSI signals are high-dimensional, noisy, and lack explicit spatial encoding, making the learning problem highly nonlinear. In practice, this results in poor generalization across different room configurations, substantial variance in reconstruction quality, and point clouds that are structurally inconsistent or incomplete. On the contrary, the proposed method introduces a structured and modular framework that explicitly separates geometric reasoning from signal interpretation. The PointNet autoencoder learns a robust latent representation that captures the spatial semantics and geometry regularities of indoor environments. Subsequently, by enforcing similarity between the latent vectors of the CSI encoder and those of the PointNet autoencoder, the model effectively anchors the signal-based representation to a geometrically meaningful domain. This approach introduces multiple advantages. First, it significantly reduces the complexity of the CSI-to-geometry mapping task. Instead of inferring 3D structures directly, the CSI encoder only needs to approximate a latent vector that is already known to correspond to valid and coherent point clouds. Second, the two-stage framework improves training stability. The PointNet autoencoder can leverage extensive offline training, while the CSI encoder benefits from a constrained target distribution during optimization. This not only enhances convergence properties but also reduces overfitting, especially when the amount of paired CSI-point cloud data is limited. Moreover, the proposed method demonstrates superior robustness to environmental variations in the CSI data. Since the latent space is learned from geometrically rich data, it is less sensitive to the stochastic nature of wireless signals. Fig. \ref{fig:rec_pointclouds}, shows an example of pointclouds generated from acquired CSI.
\begin{table*}[t]
    \centering
    \caption{Quantitative results of the ablation study comparing the proposed method with two alternative model variants: a direct CSI-to-point cloud regression baseline and an autoencoder-based approach without latent space alignment. The proposed method achieves the lowest CD and EMD, demonstrating the effectiveness of latent space alignment in enhancing reconstruction accuracy and consistency.}
    \begin{tabular}{lcccc}
        \hline
        \hline
        \textbf{Model Variant} & \textbf{Mean CD $\downarrow$} & \textbf{Std Dev} & \textbf{Mean EMD $\downarrow$} & \textbf{Std Dev} \\
         Direct CSI-to-PointCloud Regression (No AE)  & 0.325  & 0.420 & 0.412  & 0.510  \\
        Autoencoder w/o Latent Alignment   & 0.138  & 0.071 & 0.264  & 0.058  \\
        Proposed method & \textbf{0.083}  & 0.029 & \textbf{0.175}  & 0.033  \\
        \hline
        \hline
    \end{tabular}
    \label{tab:ablation}
\end{table*}
\subsection{Ablation Study}
To further validate the individual contributions of each component in the proposed framework, we conducted an ablation study that considers the proposed model variant consisting in PointNet only. This variant involves using the PointNet autoencoder for point cloud learning, but the CSI encoder is trained independently, without any explicit latent space alignment. Here, the final layer of the CSI encoder is directly connected to the decoder. The results obtained with the variant model shows that introducing a PointNet autoencoder improves performance, indicating that even without explicit latent alignment, leveraging a learned decoder trained on structured data provides benefits. Tab, \ref{tab:ablation} reports the comparisons among the three models.

\section{Conclusion}
\label{sec:conclusion}
This paper introduced a novel deep learning framework for reconstructing 3D point clouds from WiFi Channel State Information by leveraging a two-stage architecture that aligns the latent spaces of geometric and signal representations. Through preliminary experiments, we demonstrated that the proposed method significantly outperforms a direct regression baseline, achieving lower reconstruction errors and producing more coherent and structured point clouds. The use of a PointNet autoencoder as a geometric prior, combined with a CSI encoder trained to map into its latent space, proved essential for effective cross-modal translation. These findings highlight the value of modular architectures and latent space alignment in addressing the challenges of cross-domain inference. Future work will explore scaling the framework to dynamic environments and extending it to temporal CSI sequences for continuous 3D scene understanding.

\section*{Acknowledgment}
``EYE-FI.AI: going bEYond computEr vision paradigm using wi-FI signals in AI systems'' project of the Italian Ministry of Universities and Research (MUR) within the PRIN 2022 Program (CUP: B53D23012950001).
 
\bibliographystyle{ieeetr}
\bibliography{references}

\begin{thebibliography}{10}

\bibitem{zhao2018through}
M.~Zhao, T.~Li, M.~A. Alsheikh, Y.~Tian, H.~Zhao, A.~Torralba, and D.~Katabi,
  ``Through-wall human pose estimation using radio signals,'' in {\em 2018
  IEEE/CVF Conference on Computer Vision and Pattern Recognition},
  pp.~7356--7365, 2018.

\bibitem{adib2015multi}
F.~Adib, Z.~Kabelac, and D.~Katabi, ``Multi-person localization via rf body
  reflections,'' in {\em Proceedings of the 12th USENIX Conference on Networked
  Systems Design and Implementation}, p.~279–292, USENIX Association, 2015.

\bibitem{hasmath2020device}
H.~F. {Thariq Ahmed}, H.~Ahmad, and A.~C.V., ``Device free human gesture
  recognition using wi-fi csi: A survey,'' {\em Engineering Applications of
  Artificial Intelligence}, vol.~87, p.~103281, 2020.

\bibitem{wang2014eyes}
Y.~Wang, J.~Liu, Y.~Chen, M.~Gruteser, J.~Yang, and H.~Liu, ``E-eyes:
  device-free location-oriented activity identification using fine-grained wifi
  signatures,'' in {\em Proceedings of the 20th Annual International Conference
  on Mobile Computing and Networking}, MobiCom '14, p.~617–628, Association
  for Computing Machinery, 2014.

\bibitem{adib2013see}
F.~Adib and D.~Katabi, ``See through walls with wifi!,'' {\em SIGCOMM Comput.
  Commun. Rev.}, vol.~43, no.~4, p.~75–86, 2013.

\bibitem{donarski2020environment}
A.~Donarski, I.~Collings, and S.~Hanly, ``Environment mapping using wireless
  channel state information and deep learning,'' in {\em 2020 14th
  International Conference on Signal Processing and Communication Systems
  (ICSPCS)}, 2020.

\bibitem{charles2017pointnet}
R.~Q. Charles, H.~Su, M.~Kaichun, and L.~J. Guibas, ``Pointnet: Deep learning
  on point sets for 3d classification and segmentation,'' in {\em 2017 IEEE
  Conference on Computer Vision and Pattern Recognition (CVPR)}, pp.~77--85,
  2017.

\bibitem{zhou2018voxelnet}
Y.~Zhou and O.~Tuzel, ``{ VoxelNet: End-to-End Learning for Point Cloud Based
  3D Object Detection },'' in {\em 2018 IEEE/CVF Conference on Computer Vision
  and Pattern Recognition (CVPR)}, (Los Alamitos, CA, USA), pp.~4490--4499,
  IEEE Computer Society, 2018.

\bibitem{gupta2015aligning}
S.~Gupta, P.~Arbeláez, R.~Girshick, and J.~Malik, ``Aligning 3d models to
  rgb-d images of cluttered scenes,'' in {\em 2015 IEEE Conference on Computer
  Vision and Pattern Recognition (CVPR)}, pp.~4731--4740, 2015.

\bibitem{lecun2015deep}
Y.~LeCun, Y.~Bengio, and G.~Hinton, ``Deep learning,'' {\em Nature}, vol.~521,
  no.~7553, pp.~436--444, 2015.

\bibitem{agarap2019deeplearningusingrectified}
A.~F. Agarap, ``Deep learning using rectified linear units (relu),'' {\em
  arXiv}, no.~1803.08375, 2019.

\bibitem{guo2024lidarnet}
Y.~Guo, Y.~Li, D.~Ren, X.~Zhang, J.~Li, L.~Pu, C.~Ma, X.~Zhan, J.~Guo, M.~Wei,
  Y.~Zhang, P.~Yu, S.~Yang, D.~Ji, H.~Ye, H.~Sun, Y.~Liu, Y.~Chen, J.~Zhu, and
  H.~Liu, ``Lidar-net: A real-scanned 3d point cloud dataset for indoor
  scenes,'' in {\em Proceedings of the IEEE/CVF Conference on Computer Vision
  and Pattern Recognition (CVPR)}, 2024.

\bibitem{schoenberger2016sfm}
J.~L. Sch\"{o}nberger and J.-M. Frahm, ``Structure-from-motion revisited,'' in
  {\em Conference on Computer Vision and Pattern Recognition (CVPR)}, 2016.

\end{thebibliography}

\end{document}